\documentclass[letterpaper,journal]{IEEEtran}
\usepackage{graphicx}
\usepackage{cite}       
\graphicspath{ {./images/} }
\usepackage{xcolor}
\usepackage{amsmath,amsfonts,amssymb}
\usepackage{siunitx}
\usepackage{multirow}
\usepackage{tabularray}
\usepackage{pgfplots}
\usetikzlibrary{patterns}
\pgfplotsset{compat=1.15}
\usepgfplotslibrary{statistics}
\usepackage{enumitem}  

\newcommand{\reals}{{\mbox{\bf R}}}

\newcommand{\sign}{\mathop{\textnormal{sign}}}

\newcommand{\eg}{{\it e.g.}}
\newcommand{\ie}{{\it i.e.}}

\newcommand{\bs}[1]{\boldsymbol{#1}} 
\newcommand{\kneeSOMAstiffness}{\SI{12.92}{\newton\meter\per\radian}}
\newcommand{\kneeSOMApreload}{\SI{0.21}{\newton\meter}}
\newcommand{\kneeSOMATorquePercent}{\SI{24}{\percent}}
\newcommand{\kneeSOSAstiffness}{\SI{6.37}{\newton\meter\per\radian}}
\newcommand{\kneeSOSApreload}{\SI{-10.72}{\newton\meter}}
\newcommand{\kneeSOSATorquePercent}{\SI{26.1}{\percent}}   
\newcommand{\ankleSOMAstiffness}{\SI{2.9}{\newton\meter\per\radian}}
\newcommand{\ankleSOMApreload}{\SI{0.51}{\newton\meter}}
\newcommand{\ankleSOSAstiffness}{\SI{ 5}{\newton\meter\per\radian}}
\newcommand{\ankleSOSApreload}{\SI{0.62}{\newton\meter}}

\begin{document}
\title{Extending the Benefits of Parallel Elasticity across Multiple Actuation Tasks: A Geometric and Optimization-Based Approach}
\author{Kang Yang, Myia Dickens, James Schmiedeler, and Edgar Bol\'ivar-Nieto
\thanks{This material is based upon work supported by the U.S. National Science Foundation under Grant No. 2344765 and 2344766. Any opinions, findings, and conclusions or recommendations expressed in this material are those of the authors and do not necessarily reflect the views of the National Science Foundation. All the authors are with the Department of Aerospace and Mechanical Engineering, University of Notre Dame, Notre Dame, IN 46556, USA. Emails:{\tt\small\{kyang3, mdicken2, James.P.Schmiedeler.4, ebolivar\}@nd.edu}}
}
\markboth{IEEE/ASME Transactions on Mechatronics}{N/A}
\maketitle
\begin{abstract}
A spring in parallel with an effort source (\eg, electric motor or human muscle) can reduce its energy consumption and effort (\ie, torque or force) depending on the spring stiffness, spring preload, and actuation task. However, selecting the spring stiffness and preload that guarantees effort or energy reduction for an arbitrary set of tasks is a design challenge. This work formulates a convex optimization problem to guarantee that a parallel spring reduces the root-mean-square source effort or energy consumption for multiple tasks. Specifically, we guarantee the benefits across multiple tasks by enforcing a set of convex quadratic constraints in our optimization variables, the parallel spring stiffness and preload. These quadratic constraints are equivalent to ellipses in the stiffness and preload plane; any combination of stiffness and preload inside the ellipse represents a parallel spring that minimizes effort source or energy consumption with respect to an actuator without a spring. This geometric interpretation intuitively guides the stiffness and preload selection process. We analytically and experimentally prove the convex quadratic function of the spring stiffness and preload. As applications, we analyze the stiffness and preload selection of a parallel spring for a knee exoskeleton using human muscle as the effort source and a prosthetic ankle powered by electric motors. The source code associated with our framework is available as supplemental open-source software.
\end{abstract}
\begin{IEEEkeywords}
Parallel Elastic Actuators \& Wearable Robotics.
\end{IEEEkeywords}

\section*{Selected Nomenclature}
\begin{description}[labelwidth = 3.5em]
    \item[$\reals, \reals^n$] Set of real numbers and $n$-vectors ($n\times 1$ matrices).
    \item[$\reals^{m \times n}$] Set of real $m \times n$ matrices.
    \item[$\tau_{\{s,e,l\}}$] Spring, source, or load effort (\ie, force or torque).
    \item[$\tau_{\{m,u\}}$] Electromagnetic or unmodeled torque.
    \item[$q_{\{l,m\}}$] Load or motor position.
    \item[$k_p,\ \tau_p$] Parallel spring stiffness and preload effort.
    \item[$J_m$] Mass moment of inertia (motor and transmission).
    \item[$\eta, \mu$] Viscous and Coulomb friction coefficients.
    \item[$R_m, L_m$] Winding electrical resistance and inductance.
    \item[$r$] Transmission reduction ratio.
    \item[$\bs{x}$] Optimization variable in $\reals^2$ (stiffness and preload).
    \item[$r, e$] RMS effort and energy consumption subscripts.
    \item[$\bs{Q}$] $\bs{Q}_r$ or $\bs{Q}_e$, quadratic form matrix ($\bs{x}^T\bs{Qx}$) in $\reals^{2 \times 2}$.
    \item[$\bs{q}$] $\bs{q}_r$ or $\bs{q}_e$, linear form vector ($\bs{q}^T\bs{x}$) in $\reals^{2}$.
    \item[$c$] $c_r$ or $c_e$, RMS effort or energy without spring.
\end{description}

\section{Introduction}
In robotic actuation, elastic components can be in series or parallel with an effort source to create a Parallel or Series Elastic Actuator (PEA or SEA). In this context, effort (\eg, torque or force) and flow (\eg, angular or linear velocity) characterize the mechanical power exchange between a mechanical load and the actuator \cite{hogan_impedance_1985}. Elasticity is a design consideration in applications that prioritize mass, energy efficiency, and safety of physical interaction, such as wearable robots \cite{bicchi_fast_2004,park_safe_2008,jimenez-fabian_reduction_2017,vanderborght_comparison_2009, beckerle_series_2017,verstraten_series_2016,bravo-palacios_engineering_2024}. If the effort source is an electric motor connected to a mechanical transmission, the series spring decouples the load and motor kinematics, which can reduce impact loads \cite{au_powered_2008}, modify motor power, and enable joint torque control by controlling the spring deflection \cite{pratt_series_1995,dong_design_2020}. However, the load torque transfers directly into the transmission. Thus, in a backdrivable system, a series spring has negligible influence on the motor torque and the corresponding heat losses -- exclusively affecting kinematic-related loads such as inertial and friction torques \cite{mazumdar_parallel_2017,bolivar-nieto_convex_2021,robert_brown_maneuver_2013}. 

In contrast, springs in parallel directly influence the effort of the source and have the potential to reduce it in comparison with a series- or no-spring configuration. For example, parallel springs can offload the torque required to compensate for gravity \cite{jimenez-fabian_reduction_2017,beckerle_series_2017}. The potential effort reduction depends on the actuation task, the spring stiffness, and the spring preload \cite{grimmer_comparison_2012,mazumdar_parallel_2017,yang_design_2008}. Traditionally, there are two methods to select the parallel spring stiffness and preload: \textit{numerical optimization} and \textit{passive dynamics}.

In \textit{numerical optimization}, the designer formulates an optimization objective (\eg, energy consumption) and constraints as functions of the optimization variables that parameterize the load-elongation curve (\eg, the coefficients of a polynomial) \cite{realmuto_nonlinear_2015}. Without a particular structure, the numerical optimization approach may be sensitive to initial conditions, may not converge, and does not guarantee that the resulting load-elongation profile is the absolute best. Finding structure in the optimization problem (\ie, convexity) eliminates these drawbacks \cite{bolivar_nieto_minimizing_2019,bolivar-nieto_convex_2021,bolivar-nieto_convex_2021-1}. In \textit{passive dynamics}, the designer finds a load-deflection spring profile that passively generates the load kinetic and kinematic requirements as much as possible. This process can rely on a plot of the required torque and position to find a spring-deflection profile that passively accomplishes the actuation task. For example, Mazumdar et al. \cite{mazumdar_parallel_2017} used the passive-dynamics approach to experimentally reduce motor torque for the hip adduction joint of a biped robot in different walking gaits. The passive-dynamics approach is intuitive for experienced designers but lacks optimality guarantees and, in general, the load may not fit the torque-displacement relationship of a passive spring. A big challenge in both methods is to guarantee the benefits of parallel springs in a set of distinct tasks, such as walking, running, and going upstairs \cite{pfeifer_actuator_2015,hsieh_design_2017,gao_implementation_2019,sharbafi_parallel_2019,mazumdar_parallel_2017}.

\textbf{The contribution of this work} is a method to select the stiffness and preload of a PEA that minimizes root-mean-square (RMS) source effort or energy consumption for an arbitrary set of actuation tasks. Our formulation applies to arbitrary effort sources, such as human muscles, hydraulic cylinders, or electric motors. The method relies on optimization objectives represented as convex quadratic functions of parallel spring stiffness and preload.  Multiple quadratic functions define ellipses in the stiffness and preload plane. Points inside the ellipses guarantee a reduction of the optimization objective for multiple tasks. The convex quadratic functions also define an optimization problem to numerically calculate the optimal stiffness and preload torque for a large set of tasks. The two approaches are complementary. The intersection of ellipses is intuitive but may be difficult to scale for a high number of tasks; numerical optimization provides stiffness and preload for an arbitrary set of tasks but lacks graphical intuition. Compared to our prior work \cite{guo_convex_2022}, our contribution includes spring preload as an optimization variable, considers arbitrary effort sources, and guarantees the benefits of parallel springs for multiple actuation tasks through the combination of a geometric- and optimization-based approach.

Five sections organize the content of this work. Section~\ref{sec:DesignOptimalStiffness} describes our parallel stiffness and preload design framework, including the derivation of RMS source effort as a convex quadratic function of stiffness and preload. Section~\ref{sec:ExpValidation} experimentally validates the convex functions and applies our framework to specify the stiffness and preload for a knee exoskeleton and a powered ankle prosthesis. The knee spring minimizes human sit-to-stand RMS-torque, while the ankle spring minimizes energy consumption of an electric motor. Sections~\ref{sec:discussion} and \ref{sec:conclusion} discuss and conclude this work.
\section{Methods: a parallel stiffness and preload design framework}\label{sec:DesignOptimalStiffness}
\subsection{Effort as an affine function of stiffness and preload}\label{sec:EffortAsAffine}
The convexity of RMS source effort depends on the affine relationship between effort and our optimization variables, the parallel spring stiffness and preload. This section introduces the affine relationship for an arbitrary effort source and then for an electric motor coupled with a mechanical transmission.
\subsubsection{Generic effort source}\label{sec:GenericEffortSourceAsAffine}
The balance of torques or forces between the PEA and the load (Newton's third law) states
\begin{align}
    \tau_s(t) + \tau_e(t) + \tau_l(t) = 0,\label{eq:EffortBalance}
\end{align}
where $t$ is time and $\tau_s(t)$, $\tau_e(t)$, $\tau_l(t)$ are the effort of the spring, the effort source, and the load (Fig.~\ref{fig:PEA_diagram}). The spring effort is a linear function of spring stiffness and preload, 
\begin{align}
    \tau_s(t) = - k_p q_l(t) + \tau_p, \label{eq:SpringTorque}
\end{align}
where $q_l(t)$ is the load position trajectory, and $k_p,\ \tau_p \in \reals$ are the spring stiffness and its preload effort. The load kinematics and kinetics can be continuous- or discrete-time trajectories. However, the rest of this paper will omit $t$ to simplify the notation. This work assumes that $q_l$, $\dot{q}_l$, $\ddot{q}_l$, and $\tau_l$ are known inputs to the spring design process. Substituting (\ref{eq:SpringTorque}) into (\ref{eq:EffortBalance}), the source effort results in the following affine function of the parallel stiffness and preload:
\begin{align}
    \tau_e & = k_p q_l - \tau_p - \tau_l, \nonumber \\
    & = \bs{a}_e^T \bs{x} + b_e,
    \label{eq:EffortAffine}
\end{align}
where
\begin{align*}
    \bs{x} &= 
    \begin{bmatrix}
    k_p  \\
    \tau_p 
    \end{bmatrix},\ \bs{a}_e = 
    \begin{bmatrix}
    q_l \\
    - 1
    \end{bmatrix},\ \textnormal{and } b_e = - \tau_l.    
\end{align*}

\subsubsection{Electric motor as an effort source}\label{sec:motorTorqueAsEffortSoure}
\begin{figure}
  \begin{center}
  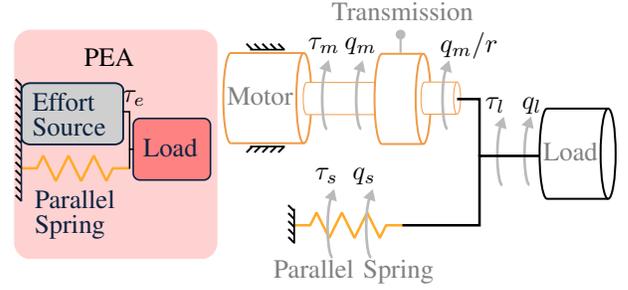
  \end{center}
  \caption{In this work, a PEA denotes an effort source (\eg, human muscle or hydraulic cylinder) connected in parallel with a spring to actuate a load (left). The figure on the right defines the coordinates when the effort source is an electric motor connected to a mechanical transmission.}
  \label{fig:PEA_diagram}
\end{figure}
Using an electric motor as an effort source, the effort becomes the motor torque. In this case, the PEA refers to an electric motor, a transmission, and a parallel spring (Fig.~\ref{fig:PEA_diagram}). Using the Newton-Euler equations, the balance of torques between the motor and transmission, (\ref{eq:EffortBalance}), states
\begin{align}
    \tau_m &= J_m\ddot{q}_m + \eta\dot{q}_m + \mu \sign(\dot{q}_m) + \tau_u - \frac{\tau_l+\tau_s}{r}, \label{eq:motorTorque} 
\end{align}
where $J_m$ is the combined motor and transmission mass-moment-of-inertia; $\eta$ is the viscous friction coefficient from the motor, the transmission, and the spring; $\mu$ is the torque due to Coulomb friction in the transmission; $r$ is the reduction ratio; $q_m$, $\dot{q}_m$, and $\ddot{q}_m$ are the motor position, velocity, and acceleration; and $\tau_m$ and $\tau_u$ are the motor electromagnetic torque and unmodeled torques (\eg, cogging torque or any torque that can be expressed as a function of the motor kinematics) \cite{verstraten_modeling_2015}. The torque losses from the transmission (\ie, Coulomb and viscous friction) use the motor kinematics due to the rigid connection between the motor and the gearbox shaft \cite{verstraten_series_2016}. The position, velocity, and acceleration of the spring and load are equal; they differ from the motor speed only by the reduction ratio ($r$), \ie,
\begin{equation}
    \frac{q_m}{r} = q_s =q_l, \; \frac{\dot{q}_m}{r} = \dot{q}_s = \dot{q}_l, \; \frac{\ddot{q}_m}{r} = \ddot{q}_s = \ddot{q}_l.
    \label{eq:KinematicsRelation}
\end{equation}
Substituting (\ref{eq:SpringTorque}) and (\ref{eq:KinematicsRelation}) into (\ref{eq:motorTorque}), we obtain motor torque as an affine function of spring stiffness and preload,
\begin{align}
    \tau_m & = J_m\ddot{q}_m + \eta\dot{q}_m + \mu \sign(\dot{q}_m) + \tau_u - \frac{\tau_l - k_p q_l + \tau_p}{r}, \nonumber \\
    & = \bs{a}_m^T \bs{x} + b_m,
    \label{eq:TorqueAffine}
\end{align}   
where
\begin{align*}
    \bs{a}_m &= \frac{1}{r}
    \begin{bmatrix}
    q_l \\
    - 1
    \end{bmatrix},\\ 
    b_m & = J_m\ddot{q}_lr + \eta\dot{q}_lr + \mu \sign(\dot{q}_lr) - \tau_l r^{-1} + \tau_u.
\end{align*}
To simplify notation, we will refer to 
\begin{align}
    \tau = \bs{a}^T \bs{x} + b \label{eq:EfforOrTorqueAsAffine}
\end{align}
as the affine expression of $\bs{x}$ that defines motor torque or source effort. The values of $\bs{a}$ and $b$ come from the equivalent parameters in (\ref{eq:EffortAffine}) or (\ref{eq:TorqueAffine}).
\subsection{Convex Quadratic Functions of Stiffness and Preload}
\subsubsection{RMS source effort}\label{sec:RMSEffortAsQuadratic}
The RMS effort summarizes in a scalar value the typical magnitude of an effort trajectory over time. For example, a rule of thumb to prevent overheating is to select a motor with a rated continuous torque higher than the RMS load torque. This section formulates a convex quadratic function of $\bs{x}$ that minimizes RMS source effort,
\begin{equation*}
   \text{RMS}(\tau)  = \sqrt{\dfrac{1}{t_f-t_0}\int_{t_0}^{t_f} \tau(t)^2 dt}.
\end{equation*}
In this work, we minimize $\text{RMS}(\tau)$ by minimizing a function that excludes the square root and the positive scalar $(t_f - t_0)^{-1}$. The square root is a monotonically increasing function over nonnegative reals and does not affect the value of $\bs{x}$ that minimizes $\text{RMS}(\tau)$. Thus, substituting (\ref{eq:EfforOrTorqueAsAffine}), we minimize $\text{RMS}(\tau)$ by minimizing the function
\begin{align}
   \int_{t_0}^{t_f} \tau(t)^2 dt  &= \int_{t_0}^{t_f}  \left( \bs{x}^T\bs{a}\bs{a}^T\bs{x} + 2 \bs{a}^T b \bs{x} + b^2 \right)dt, \nonumber \\
    &= \bs{x}^T \bs{Q}_r \bs{x} + 2 \bs{q}_r^T \bs{x} + c_r, \label{eq:RMSTorqueEquivalent}
\end{align}
where
\begin{flalign*}
    \bs{Q}_r &= \int_{t_0}^{t_f} {\bs{a} \bs{a}^T}\ dt, & \bs{q}_r &= \int_{t_0}^{t_f} \bs{a}b\ dt, & c_r &= \int_{t_0}^{t_f}  b^2\ dt.
\end{flalign*}
The quadratic function (\ref{eq:RMSTorqueEquivalent}) is convex with respect to $\bs{x}$ because its second-order derivative ($2\bs{Q}_r$) is a positive semidefinite matrix \cite{boyd_convex_2004}---$\bs{Q}_r$ is the integral (positive weighted sum, \ie, $t_f > t_0$) of the matrix $\bs{a}\bs{a}^T$, which has nonnegative eigenvalues. For example, with $\bs{a}$ as defined in (\ref{eq:TorqueAffine}) the two eigenvalues of $\bs{a}\bs{a}^T$ are $\lambda_1 = 0 \text{ and } \lambda_2 = (1+q_l^2)r^{-2}$. 
\subsubsection{Motor energy losses} \label{sec:ConvexFormulationEnergyLosses}
If the effort source is an electric motor, it is possible to use the affine expression (\ref{eq:TorqueAffine}) to minimize the motor energy losses using convex quadratic functions of the parallel spring stiffness and preload. Using Kirchhoff’s voltage law across the motor’s winding, we model the electrical behavior of the PEA’s motor as
\begin{align}
    v_s & = i_m R_m + L_m\frac{di_m}{dt} + v_\textrm{emf}, \nonumber \\
    & \approx i_m R_m + v_\textrm{emf},
    \label{eq:supplyVoltage}
\end{align}
where $v_s$ is the power supply voltage, $i_m$ the motor current, $R_m$ the winding resistance, $L_m$ the winding inductance, and $v_\textrm{emf}$ the motor electromotive voltage. We assume that the voltage drop due to the winding inductance is negligible compared to the voltage drop due to the winding resistance.

For a DC brushed motor, its electromagnetic torque and electromotive voltage are linear functions of the current and motor speed,
\begin{equation}
    \tau_m = k_t i_m,\  v_\textrm{emf} = k_v \dot{q}_m,\ k_m = k_t/\sqrt{R_m}, \label{eq:torqueEMF}
\end{equation}
where $k_t$, $k_v$, and $k_m$ are the torque, voltage, and motor constants, respectively. In SI-units, $k_t$ = $k_v$ \cite{hollerbach_comparative_1992}. The model in (\ref{eq:supplyVoltage}) applies to permanent-magnet synchronous motors when expressing the three-phase currents as a single q-axis current using the Clarke-Park transform \cite{lee_empirical_2019,park_two-reaction_1929}.

The energy consumption from the power source is
\begin{equation}
    E_m = \int_{t_0}^{t_f} i_m v_s \,dt.
    \label{eq:electricalEng}
\end{equation}
Substituting (\ref{eq:supplyVoltage}) and (\ref{eq:torqueEMF}) into (\ref{eq:electricalEng}), this energy consumption splits into dissipated winding Joule heating and the mechanical energy provided or dissipated by the rotor \cite{verstraten_modeling_2015,bolivar_nieto_minimizing_2019}.
\begin{equation}
    E_m = \int_{t_0}^{t_f} \biggl( \underbrace{ \frac{\tau_m^2}{k_m^2}}_{\substack{\mathrm{Winding}\\\mathrm{Joule\ heating}}} + \underbrace{ \tau_m \dot{q}_m }_{\substack{\mathrm{Rotor}\\\mathrm{mechanical\ power}}} \biggr) \, dt.
    \label{eq:energyEqSimple}
\end{equation}
Using the affine function of motor torque (\ref{eq:TorqueAffine}), we can write energy consumption as a quadratic function of the parallel spring stiffness and preload,
\begin{align}
    E_m & = \int_{t_0}^{t_f} \biggl[ \frac{\bs{x}^T\bs{a}\bs{a}^T\bs{x} + 2 \bs{a}^T b \bs{x} + b^2}{k_m^2} + ( \bs{a}^T \bs{x} + b ) \dot{q}_m \biggr] \,dt, \nonumber \\
    & = \bs{x}^T \bs{Q}_e \bs{x} + 2 \bs{q}_e^T \bs{x} + c_e,
    \label{eq:energyEq}
\end{align}
where
\begin{align*}
    \bs{Q}_e &= \int_{t_0}^{t_f} \frac{\bs{a} \bs{a}^T} {k_m^2}\ dt,\hspace{1cm} \bs{q}_e = \int_{t_0}^{t_f} \biggl( \frac{b}{k_m^2} + \frac{\dot{q}_lr}{2} \biggr)\bs{a}\ dt,\\
    c_e &= \int_{t_0}^{t_f} \left( \frac{b^2}{k_m^2} + b\dot{q}_m \right)\ dt. 
\end{align*}
This quadratic function is convex with respect to $\bs{x}$ for the same reasons that (\ref{eq:RMSTorqueEquivalent}) is convex. To simplify notation,
\begin{equation}
    f_0(\bs{x}) = \bs{x}^T \bs{Q} \bs{x} + 2 \bs{q}^T \bs{x} + c 
    \label{eq:GeneralQuadratic}
\end{equation}
refers to the $\text{RMS}(\tau)$ or the motor-energy-consumption functions. The corresponding expressions in (\ref{eq:RMSTorqueEquivalent}) and (\ref{eq:energyEq}) specify the parameters $\bs{Q}$, $\bs{q}$, and $c$.
\subsection{Extending the benefits of elasticity across multiple tasks} \label{sec:MultiConvex}
\subsubsection{The geometric approach}
In our formulation, a parallel spring is beneficial if it reduces (\ref{eq:GeneralQuadratic}) compared to the configuration without a spring, which is equivalent to evaluating (\ref{eq:GeneralQuadratic}) with stiffness and preload equal to zero, $f_0(0) = c$. Thus, a combination of stiffness and preload is beneficial if it satisfies the inequality
\begin{align}
   \bs{x}^T \bs{Q} \bs{x} + 2 \bs{q}^T \bs{x} + c &\leq c. \nonumber
\end{align}
This inequality describes the $c$-sublevel set of a convex quadratic function, which is equivalent to the ellipse
\begin{align}
   \mathcal{E} = \{\bs{x} \in \reals^2 \mid (\bs{x} - \bs{x}_c)^T\bs{P}^{-1}(\bs{x} - \bs{x}_c) &\leq 1\}, \label{eq:Ellipse}
\end{align}
where the point $\bs{x}_c = -\bs{Q}^{-1}\bs{q}$ is the ellipse center and $\bs{P} = \bs{Q}^{-1}(\bs{q}^T\bs{Q}^{-1}\bs{q})$ the ellipse semiaxes \cite{boyd_convex_2004}. This ellipse in the stiffness and preload plane is unique for a given load kinematics and kinetics trajectory, \ie, a given actuation task. Thus, a parallel spring is beneficial for an arbitrary set of tasks if it belongs to the intersection of their corresponding ellipses. This intersection has two properties:
\begin{enumerate}
    \renewcommand{\theenumi}{\alph{enumi}}
    \item The intersection of an arbitrary set of tasks has at least one common point, $\bs{x} = 0$, which describes the trivial case without a parallel spring.
    \item When the objective is to minimize RMS motor torque, the ellipses are independent of the electrical motor parameters (as seen from the definition of $\bs{Q}_r$, $\bs{q}_r$, and $c_r$). Thus, designers can find regions in the stiffness and preload plane that are agnostic to the selection of motor electrical properties (\eg, torque and motor constants).
\end{enumerate}

\subsubsection{The optimization-based approach}
The geometric approach provides visual intuition in the design process; however, finding the intersection of ellipses for a large number of tasks can be difficult to scale. Another approach is to numerically solve the convex Quadratically Constrained Quadratic Program (QCQP) \cite{boyd_convex_2004}
\begin{align}
\text{minimize} \quad & \bs{x}^T \bs{Q} \bs{x} + 2 \bs{q}^T \bs{x} + c; \label{eq:QCQP}\\
\textrm{subject to} \quad & \bs{x}^T \bs{Q}_i \bs{x} + 2 \bs{q}_i^T \bs{x} \leq 0; \quad i = 1,\hdots,m; \nonumber
\end{align}
where the objective function describes the main actuation task and the $m$ constraints represent tasks for which the parallel spring must be beneficial. In this case, the parallel spring is optimal for one actuation task and remains beneficial for a set of $m$ tasks. The objective and constraint functions can interchange depending on the design goals. For example, the parallel spring can be optimal for multiple tasks by writing the constraint functions as part of the optimization objective (\eg, as a weighted sum of multiple quadratic functions as shown in Section-\ref{sec:DesignProstheticAnkle}).

\subsection{Optimization constraints: guaranteeing feasibility}
An electric motor is feasible for a particular actuation task if the load torque and speed are within the operational region of the motor speed and torque plane. Springs in parallel with a motor influence feasibility by modifying motor torque, but they do not affect motor velocity. Parallel springs that minimize RMS motor torque or energy losses guarantee an overall torque reduction; however, they may increase torque in speed regions that would make the motor infeasible. Thus, our optimization problem can include inequality constraints to guarantee motor feasibility, \eg,
\begin{align}
    \|\tau_m\|_\infty &\leq \tau_\text{max}, \label{eq:peakTorque} \\
    \tau_m &\leq \frac{k_t}{R_m} v_s - \frac{k_t^2}{R_m} \dot{q}_m. \label{eq:MotorVelocityConstraint}
\end{align}
The inequality (\ref{eq:peakTorque}) indicates that the peak motor torque must be lower than $\tau_\text{max}$, a known value of maximum motor torque that is typically limited by the maximum current provided by the motor driver. The inequality (\ref{eq:MotorVelocityConstraint}) refers to the torque-speed relationship of electric motors (feasible motor torque decreases linearly as motor speed increases \cite{hollerbach_comparative_1992}). In these inequalities, the optimization variable appears by expressing motor torque on the left-hand side as the affine function (\ref{eq:TorqueAffine}). Thus, we lump (\ref{eq:peakTorque}) and (\ref{eq:MotorVelocityConstraint}) together using the inequality
\begin{equation*}
    \bs{Mx} \preccurlyeq \bs{p},
\end{equation*}
where $\bs{M}$ and $\bs{p}$ are optimization parameters that depend on the power supply voltage and motor characteristics (\eg, torque constant). The symbol $\preccurlyeq$ denotes element-wise inequality. Our previous work \cite{guo_convex_2022} defines $\bs{M}$ and $\bs{p}$ in detail.

The following section applies our optimization-based and geometric approaches to the design of wearable robots. In this application, an \emph{actuation task} refers to the load kinematics and kinetics corresponding to an activity of daily living (\eg, walking). Thus, we define the Spring Optimized for Multiple Activities (SOMA) as the solution to
\begin{align}
\text{minimize} \quad & \bs{x}^T \bs{Q} \bs{x} + 2 \bs{q}^T \bs{x} + c, \nonumber\\
\textrm{subject to} \quad & \bs{x}^T \bs{Q}_i \bs{x} + 2 \bs{q}_i^T \bs{x} \leq 0, \quad i = 1,\hdots,m, \label{eq:SOMA} \\
& \bs{Mx} \preccurlyeq \bs{p}, \nonumber
\end{align}
where the quadratic constraints can be included in the optimization objective, \eg, in a weighted sum. A Spring Optimized for a Single Activity (SOSA) is the solution to (\ref{eq:SOMA}) without the quadratic inequality constraints.

\section{Experimental Validation and Applications: A knee exoskeleton and a prosthetic ankle} \label{sec:ExpValidation}
This section describes the experiments that illustrate the convex relationship between RMS torque and the parallel spring design parameters (\ie, stiffness and preload). Additionally, we apply our geometric- and optimization-based approach to specify the stiffness and preload for two applications: a mechanically passive knee exoskeleton and a powered ankle prosthesis. The knee exoskeleton consists of a spring in parallel with a biological knee that minimizes sit-to-stand RMS human torque without increasing it during walking and stair ascent/descent.  In the ankle prosthesis, a spring operates in parallel with a quasi-direct drive electric motor to minimize energy consumption instead of RMS torque. This design reduces energy consumption for a weighted sum of activities (\eg, walking, stair ascent/descent, and sit-to-stand transitions), with the weighting factors determined by the relative practice for each activity within a day \cite{hongu_promoting_2019,dall_frequency_2010}. The two applications define the load position and torque from prerecorded kinematics and kinetics available in the literature.  We compare our SOMA approach against the traditional SOSA in these applications.  The results in this work used MOSEK to numerically solve (\ref{eq:SOMA}) \cite{mosek_aps_mosek_2024}. However, our initial optimization prototypes used CVX, a MATLAB-based modeling system for convex optimization \cite{grant_cvx_2020,blondel_graph_2008}. While the exoskeleton application is a simulation case study, the ankle application and convexity validation rely on experimental motor torques. These torques were measured using a dynamometer equipped with interchangeable torsional springs, which is described in the following section.
\subsection{Testbed: A dynamometer with custom torsional springs}\label{sec:ValidationUsingDynamometer}
\subsubsection{Dynamometer} 
\begin{figure}
  \centering
  \vspace*{0.1in}
  \includegraphics[width=\columnwidth]{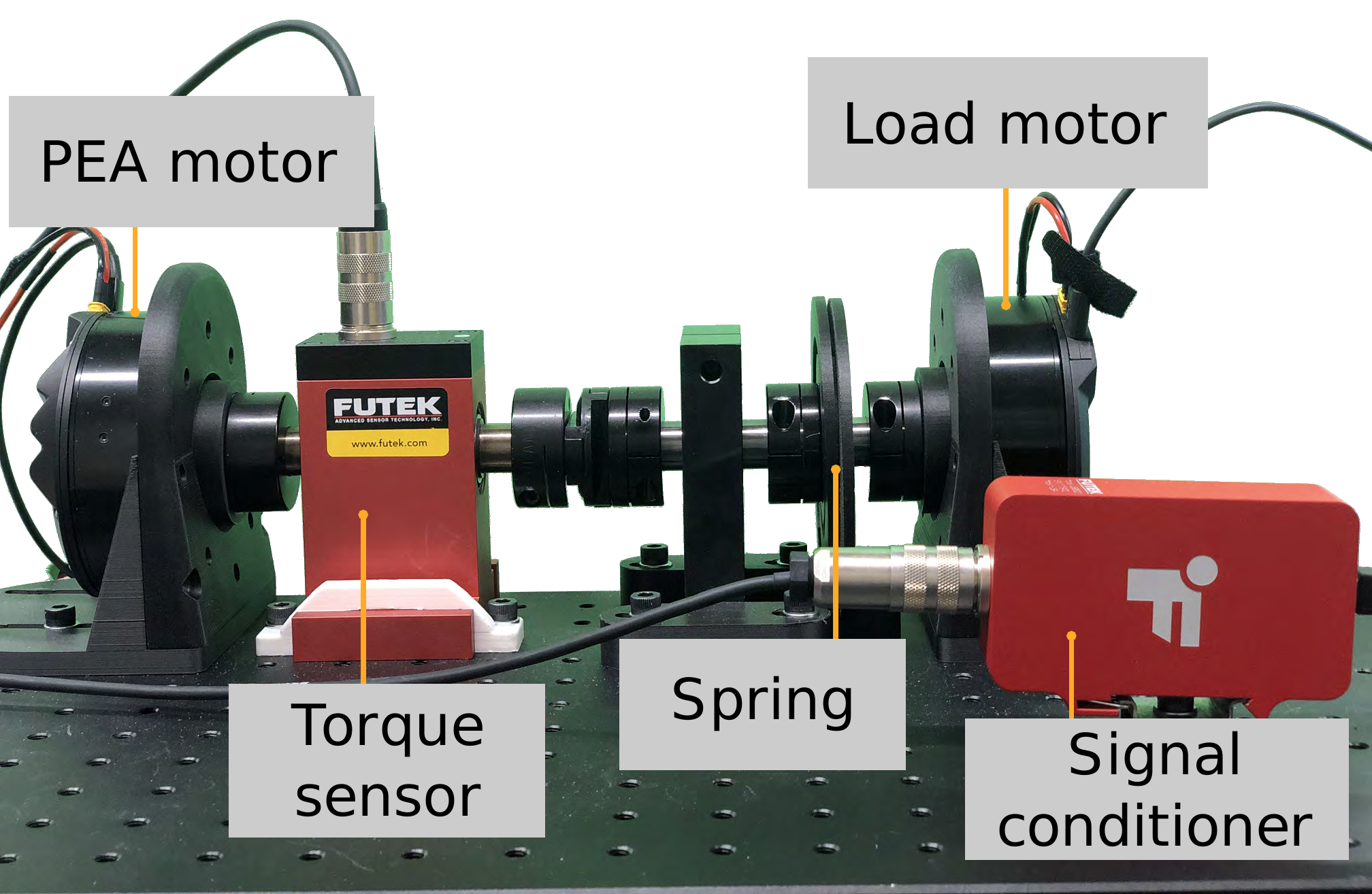}
  \caption{The testbed used for system identification and validation can simulate the dynamics of a prosthetic ankle under loading.}
  \label{fig:testbed}
\end{figure}

The experimental testbed consists of two permanent-magnet synchronous motors equipped with 9:1-reduction-ratio planetary gearboxes (ActPack 4.1, Dephy Inc., MA, USA, Table~\ref{table:MotorParam}), a custom parallel torsional spring, and a rotary torque sensor (TRS705, FUTEK, CA, USA) (Fig. \ref{fig:testbed}). A 14-bit per revolution (\SI{0.022}{\degree} resolution) absolute magnetic encoder (AK7452, Asahi Kasei, Tokyo, Japan), integrated into the ActPack, measured the rotor angle. One of the motors (load motor) emulates the load torque, while the other motor (PEA motor) tracks the load position. A Python interface sampled all the sensors and provided reference position and current commands to the ActPacks at \SI{100}{\hertz}. Energy consumption is calculated from the measured PEA motor current and velocity by using (\ref{eq:supplyVoltage}), (\ref{eq:torqueEMF}), and (\ref{eq:electricalEng}). System identification was performed using least squares, a chirp reference position signal, and the dynamic model (\ref{eq:motorTorque}). The inertia, Coulomb friction, and viscous friction of the motor, gearbox, and the spring are lumped together as they share the same position (\ie, the motor and gearbox shafts are rigidly attached to the spring). The results of system identification are in Table~\ref{table:MotorParam}.
\begin{table}[t]
    \centering
    \caption{Parameters of the ankle motor (ActPack 4.1)}
    \begin{tabular}{ p{4.5cm} p{1.5cm} }
    \hline
    Parameter & Value  \\
    \hline
    Motor constant, $k_m$ (\SI[inter-unit-product =\cdot]{}{\newton\meter\per(\watt^2)}) & 0.0916 \\
    Torque constant, $k_t$ (\SI[inter-unit-product =\cdot]{}{\newton\meter\per\ampere}) & 0.0514 \\
    Rated continuous torque, (\SI[inter-unit-product =\cdot]{}{\newton\meter}) & 3.7\\
    Gear ratio, $r$ & 9 \\
    Motor inertia, $J_m$ (\SI{}{\kilo\gram\square\meter}) & $1.2 \times 10^{-4}$ \\
    Viscous friction, $\eta$ (\SI{}{\newton\meter\second\per\radian}) & $0.16 \times 10^{-3}$   \\
    Coulomb friction, $\mu$ (\SI{}{\newton\meter}) & 0.0324  \\
    Winding resistance (\SI{}{$\Omega$}) & 0.315  \\
    Continuous current (\SI{}{\ampere}) & 8  \\
    Maximum current (\SI{}{\ampere}) & 22  \\
    Supply voltage, $v_s$ (\SI{}{\volt}) & 48  \\
    \hline
    \end{tabular}
    \label{table:MotorParam}
\end{table}
\subsubsection{Torsional springs} \label{sec:SpringDesign}
\begin{figure}
  \centering
  \includegraphics[scale=1]{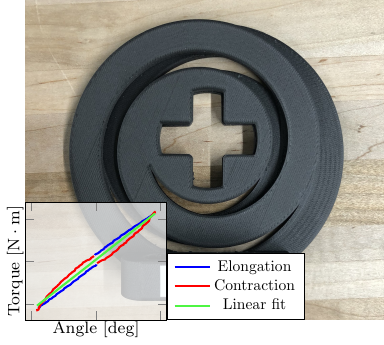}
  \caption{A 3D-printed torsional spiral spring. The double-spiral design promotes symmetric torque generation in clockwise and anticlockwise directions. The torque deflection profile is shown in the bottom left corner.}
  \label{fig:spring}
\end{figure}

The stiffness of our custom torsional springs depends on geometric and material properties (Fig.~\ref{fig:spring}),
\begin{equation*}
    k_p = \frac{Ebh^3}{12L},
\end{equation*}
where $E$ is Young’s modulus, $b$ is the width, $h$ is the thickness, and $L$ is the length \cite{munoz-guijosa_generalized_2012,mathews_design_2022}. The springs were 3D-printed using a Markforged Desktop printer with solid Onyx infill; Onyx is a composite material based on nylon filled with micro carbon fiber. Due to the asymmetric geometry of a single spiral, a single-spiral spring exhibits different stiffnesses in clockwise and counterclockwise rotations. To achieve a symmetric torque profile, we stacked two identical spirals on top of each other in a way that they are always deflected in opposite directions by the same amount, as shown in Fig. \ref{fig:spring}. The stiffness of each spring is measured by elongating the spring with a motor and reading the corresponding torque with the FUTEK torque sensor. The spring was deflected quasi-statically to avoid visco-elastic effects. We refer the reader to \cite{bons_energy-dense_2023} to design and manufacture more complex but lightweight and compact torsion springs.
%
\subsection{Experimental validation of convexity}\label{sec:ExperimentalValidationOfConvexity}
\begin{figure}
  \centering
  \includegraphics[scale = 0.99]{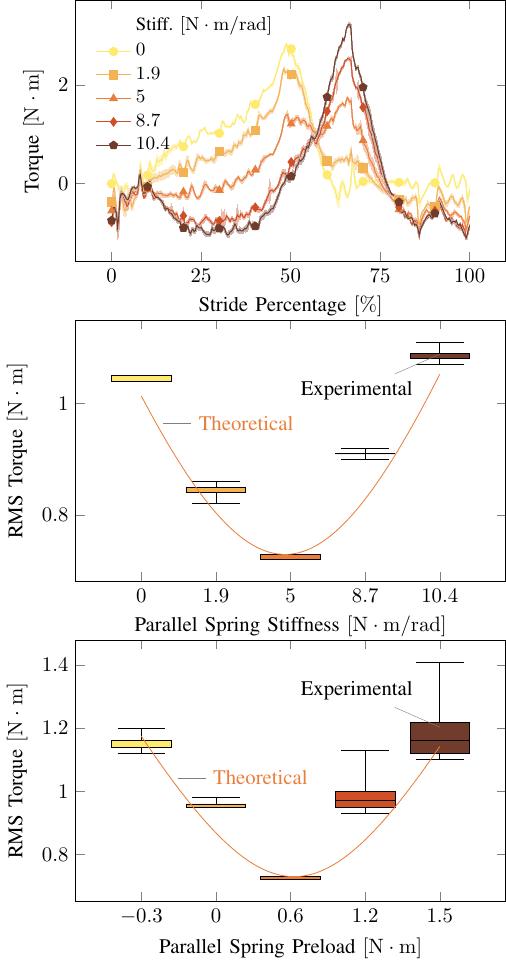}
  \caption{Convex relationship between RMS torque and the spring design parameters (\ie, stiffness and preload). The theoretical lines correspond to the quadratic functions in (\ref{eq:GeneralQuadratic}) using the load position and load torque from the biomechanics reported in Section \ref{sec:ExperimentalValidationOfConvexity} to define $\bs{Q}$, $\bs{q}$, and $c$.}
  \label{fig:ConvexRelationshipRMSTorque}
\end{figure}

To show the convexity of RMS torque as a function of stiffness, the dynamometer emulated the ankle walking kinematics and kinetics for 5 different stiffnesses with a fixed preload torque ($\tau_p = \SI{0.62}{\newton\meter}$), the optimal preload for the SOSA (see Section~\ref{sec:DesignProstheticAnkle} for details). Similarly, to show the convexity of RMS torque as a function of preload torque, the dynamometer emulated the walking kinematics and kinetics for 5 different preload torques with a fixed value of stiffness ($k_p = \SI{5.02}{\newton\meter\per\radian}$), the optimal SOSA stiffness. We approximated the ankle kinematics and kinetics in \cite{winter_biomechanical_1983} as a weighted sum of sinusoidal signals (\ie, a Fourier approximation) to enforce periodicity of our trajectories. The ankle kinetics in \cite{winter_biomechanical_1983} are normalized by body mass. We selected \SI{1.5}{\kilo\gram} as the body mass to generate load torques feasible for our actuators, with rated continuous torque of \SI{3.7}{\newton\meter} (See Table~\ref{table:MotorParam} and Section~\ref{sec:ValidationUsingDynamometer}). Selecting a body mass of \SI{70}{\kilo\gram} would require RMS motor torques around \SI{50}{\newton\meter}, which would need an additional transmission with reduction ratio 14:1, increasing the cost and mechanical complexity of our test bed. However, body mass is a scale factor that affects only the magnitude of load torque. Our conclusions will be the same regardless of this factor. Note that our actuators are the same as in the Open Source Leg (OSL) project \cite{azocar_design_2020}. The OSL achieves higher torques due to the reduction in its belt transmission \cite{azocar_design_2020}. For each stiffness and preload torque experiment, the testbed recorded PEA motor current over 50 concatenated strides to calculate the mean and standard deviation of our measured RMS torque per stride. These summary statistics address error due to uncertainty in our measurements and trajectory tracking. The experimental and theoretical RMS torques of the PEA motor with different stiffnesses and preloads illustrate the convex relationship between those variables (Fig.~\ref{fig:ConvexRelationshipRMSTorque}).

\subsection{Application I: An energy-efficient prosthetic ankle} \label{sec:DesignProstheticAnkle}
In this application, the objective of our parallel spring is to minimize energy consumption for a combination of tasks (walking, stair ascent, stair descent, and sit-stand transitions) for a prosthetic ankle. Instead of minimizing the objective for one task while guaranteeing a reduction for the others, this section exemplifies the case where the objective function $f_0(\bs{x})$ is a weighted sum of the energy consumption for each of the individual actuation tasks, $f_i(\bs{x})$, \ie,
\begin{align}
    f_0(\bs{x}) = \sum_{i=1}^4 w_if_i(\bs{x}),\hspace{0.4cm} \text{where }\hspace{0.4cm} \bs{w}=\dfrac{1}{2280}\begin{bmatrix} 2000 \\ 80 \\ 80 \\ 120 \end{bmatrix}. \label{eq:WeightedSum}
\end{align}
The percentage of daily use for each activity defines its relative weight (2000 strides for walking, 80 strides of stair ascent or descent, and 120 strides of sit-to-stand transitions \cite{hongu_promoting_2019,dall_frequency_2010}).

\subsubsection{Design from inverse kinematics and dynamics}
The ankle stiffness and preload selection uses existing biomechanical studies to define the reference trajectories (\ie, sit-to-stand \cite{roebroeck_biomechanics_1994}, stair ambulation \cite{riener_stair_2002}, and walking \cite{winter_biomechanical_1983}). In those studies, the joint torques are normalized by body mass; in \cite{roebroeck_biomechanics_1994}, they are also normalized by body height. Following the same reasoning as in Section~\ref{sec:ExperimentalValidationOfConvexity}, we assume a user mass of \SI{1.5}{\kilo\gram} and height of \SI{1.76}{\meter}, which correspond to the average height of the participants in \cite{roebroeck_biomechanics_1994} (6 females and 4 males).

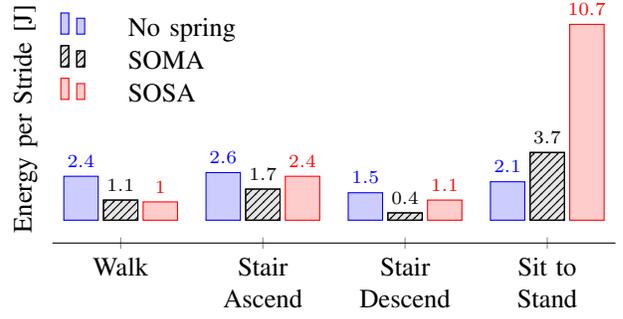
\begin{figure}
    \begin{tikzpicture}
\begin{axis}[
	width = 0.5\textwidth,
	height = 0.2\textheight,
	ybar = 2pt,
	axis x line = bottom,
	x axis line style = {-},
	axis y line = left,
	y axis line style={opacity=0},
	ytick=\empty,
	enlargelimits = 0.16,
	legend style={
		 at={(0.35,0.95)},
		draw = none,
		legend cell align = left,
		style={column sep=0.5cm},
	},
	ylabel={Energy per Stride \lbrack\SI{}{\joule}\rbrack},
	symbolic x coords={Walk, Stair Ascend, Stair Descend, Sit to Stand},
	x tick label style={
		text width=1.5cm,
		align=center,
	},
	bar width=13pt,
	xtick=data,
	nodes near coords = {\scriptsize \pgfmathprintnumber{\pgfplotspointmeta}},
	nodes near coords align={vertical},
	]
	\addplot [blue!100, fill=blue!20] coordinates {(Walk, 2.4)(Stair Ascend, 2.6)(Stair Descend, 1.5)(Sit to Stand, 2.1)};
	\addplot [black!100, fill=gray!20, postaction={pattern=north east lines}] coordinates {(Walk, 1.1)(Stair Ascend, 1.7)(Stair Descend, 0.4)(Sit to Stand, 3.7)};
	\addplot [red!100, fill=red!20] coordinates {(Walk, 1.0)(Stair Ascend, 2.4)(Stair Descend, 1.1)(Sit to Stand, 10.7)};
	\legend{No spring, SOMA, SOSA}
	\end{axis}
\end{tikzpicture}
    \caption{Motor energy consumption across multiple activities with a parallel spring optimized for a weighted sum of activities (SOMA) and a parallel spring optimized exclusively for walking (SOSA). The stiffness and preload for the parallel spring are in Fig.~\ref{fig:ellipseResultsAnkleProstheses}.}
    \label{fig:energyResultsAnkleProsthesis}
\end{figure}

\begin{figure}
    \centering
    \input{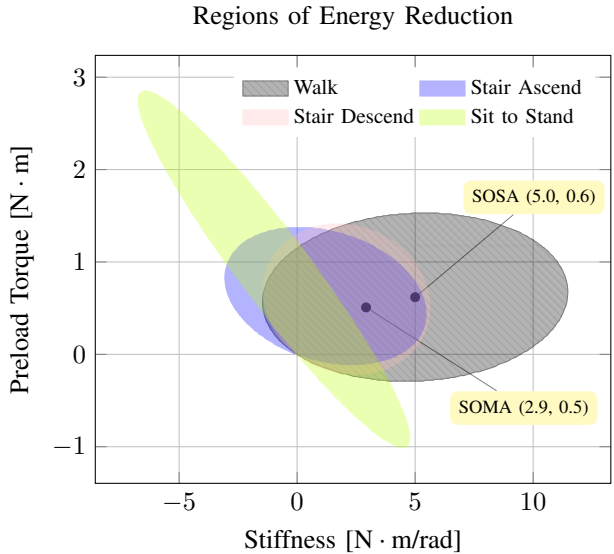}
    \caption{Stiffness and preload energy-savings regions for a spring in parallel with a motor powering a prosthetic ankle. Any point in each ellipse interior represents a combination of parallel stiffness and preload that reduces energy consumption compared to a no-spring configuration. The ellipse center denotes the optimal spring for an individual task without any consideration of other tasks (SOSA). The SOMA denotes the point that minimizes the weighted sum of activities. The SOMA (SOSA) stiffness and preload are \ankleSOMAstiffness{} (\ankleSOSAstiffness{}) and \ankleSOMApreload{} (\ankleSOSApreload{}). The semi-major axis represents a direction where changes in stiffness and preload have the smallest impact on the cost for a given ellipse (task).}
    \label{fig:ellipseResultsAnkleProstheses}
\end{figure}

Based on this data, we numerically solved (\ref{eq:SOMA}) when there are no constraints, and the objective function is defined by (\ref{eq:WeightedSum}), which corresponds to our SOMA (stiffness and preload are \ankleSOMAstiffness{} and \ankleSOMApreload{}). For comparison, we define SOSA as the spring that exclusively minimizes energy consumption during walking (stiffness and preload are \ankleSOSAstiffness{} and \ankleSOSApreload{}, which represent the center of the walking ellipse). The influence of the parallel spring on energy consumption is in Fig.~\ref{fig:energyResultsAnkleProsthesis}, and the corresponding stiffness and preload energy-saving ellipses are in Fig.~\ref{fig:ellipseResultsAnkleProstheses}. The relative weights in (\ref{eq:WeightedSum}) do not affect the shape of the ellipses, but they do affect the cost associated with each level set. One interpretation is to imagine a three dimensional plot where the third axis is the optimization cost. The weights affect the relative height of each of the elliptical paraboloids.

The level sets for each ellipse are points with equal optimization cost; thus, the ellipse orientation (given by the ellipse semiaxes or eigenvector direction) can also guide the stiffness and preload selection process. The smallest semiaxis indicates a direction where small changes in the stiffness and preload have a high impact on the cost. Thus, the long semiaxis of the walking ellipse in Fig.~\ref{fig:ellipseResultsAnkleProstheses} or the sit-to-stand ellipse in Fig.~\ref{fig:ellipseResultsKneeExo} indicate directions where the SOMA can move with respect to the SOSA while minimizing the impact on the optimization cost. Similarly, the SOSA in Fig.~\ref{fig:ellipseResultsAnkleProstheses} moves almost in parallel with the smallest semiaxis of the sit-to-stand ellipse, explaining the high penalty in energy consumption with respect to the SOMA and no-spring scenarios.
\subsubsection{Experimental Results} \label{sec:ExpResults}
The testbed described in Section~\ref{sec:ValidationUsingDynamometer} recorded the energy consumption of the PEA motor for each activity with three configurations of the parallel spring: without parallel spring, SOMA, and SOSA. The energy consumption per stride was the cumulative sum of the power consumption over a stride --- the multiplication of bus voltage and current supplied to the motor drivers. The dynamometer performed 50 strides per activity to increase the statistical power of our study. We scaled body mass to be \SI{1.5}{\kilo\gram} to specify reference load torque trajectories feasible for our actuators (See details in Section~\ref{sec:ExperimentalValidationOfConvexity}). The stiffness profile of the SOSA is shown in Fig. \ref{fig:spring}. The area inside the hysteresis loop formed by the torque deflection curves represents the amount of energy dissipated by the spring per loading cycle. Our model represents those energy losses through Coulomb and viscous friction. The stiffness of the SOSA was measured to be \SI{4.67}{\newton\meter\per\radian}, and \SI{2.70}{\newton\meter\per\radian} for the SOMA.

\begin{figure}
  \centering
  \includegraphics[width=0.48\textwidth]{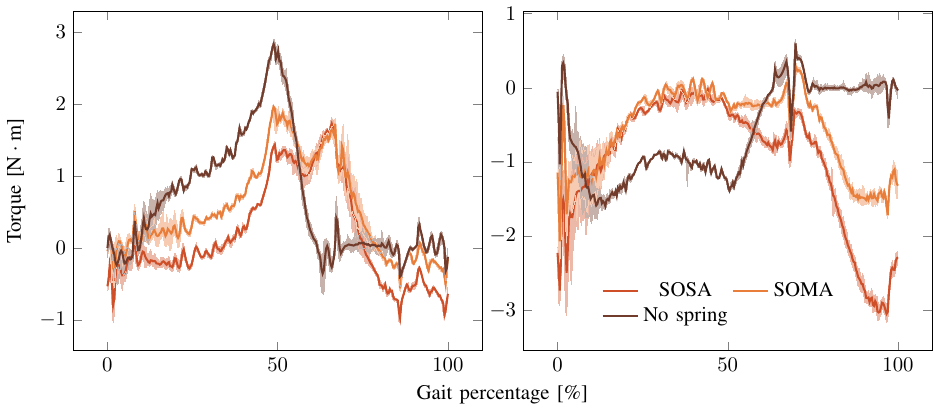}
  \caption{Experimentally measured torque of an actuator without the spring, with a spring optimized for walking (SOSA), and with a spring optimized for multiple activities (SOMA) during walking (left) and stair descent (right) for 50 strides. The solid line represents the mean value, and the area around the solid line is bounded by the minimum and maximum values.}
  \label{fig:MotorTorque}
\end{figure}

As expected, the SOSA configuration reduced walking energy consumption the most (Table~\ref{table:ExperimentalResults}), but it also increased it the most for sit-to-stand transitions. The SOMA design minimizes a weighted sum of the activities.  While it is suboptimal for walking alone, it minimizes the total energy consumption for combined use across multiple activities, as defined by the weighting factors. The motor torque trajectories for the three spring configurations illustrate the advantages of a multi-activity design (SOMA) compared to a single-activity approach (SOSA).  Specifically, the SOSA may increase peak torques in stair descent, leading to excessive heat dissipation in the motor (Fig.~\ref{fig:MotorTorque}).
\begin{table}
    \centering
    \caption{Effect of parallel spring on energy consumption}
    \begin{tabular}{ p{1.7cm} | p{0.8cm} p{0.8cm} p{0.8cm} | p{1cm} p{1cm} }
    & \multicolumn{3}{c}{Energy per stride [\SI{}{\joule}]} & \multicolumn{2}{c}{ Avg. reduction} \\
    & Without & SOMA & SOSA  & SOMA & SOSA \\
    & Spring &  &   &  &  \\
    \hline
    Walk          & 3.41   & 2.14 & 1.88 & 37.17\%   & 44.87\%\\
    Stair Ascent  & 3.84   & 2.77 & 3.47 & 27.93\%   & 9.57\%\\
    Stair Descent & 2.25   & 1.35 & 2.16 & 40.01\%   & 4.03\%\\
    Sit to Stand  & 2.53   & 4.42 & 11.2 & -74.87\%  & -342.8\%\\
    \hline
    Weighted Sum  & 3.34   & 2.26 & 2.44 & 32.34\%  & 26.9\%\\
    \end{tabular}
    \label{table:ExperimentalResults}
\end{table}
%
\subsection{Application II: A knee exoskeleton to assist sit-to-stand}\label{sec:ExoDesign}
Sit-to-stand transitions can be physically demanding for older adults and the caregivers assisting them, who are at high risk of back injury occurrence and prevalence \cite{bohannon_measurement_2012,garg_reducing_1992,garg_biomechanical_1991,garg_biomechanical_1991-1,hignett_workrelated_1996,dawson_interventions_2007}. Compared to the hip and ankle, the knee provides the highest torque to lift the whole-body center of mass during sit-to-stand transitions \cite{roebroeck_biomechanics_1994}. Thus, this section applies our design framework to specify the stiffness and preload of a spring in parallel with a human knee to reduce its sit-to-stand RMS torque without increasing the same during walking, stair ascent, and stair descent. This parallel spring can be implemented in a passive exosuit to assist older adults \cite{xiloyannis_soft_2022}.
\subsubsection{Design from inverse kinematics and dynamics}
The inputs to our design framework are the knee kinematics and torques reported from inverse kinematics and dynamics studies (\ie, sit-to-stand \cite{roebroeck_biomechanics_1994}, stair ambulation \cite{riener_stair_2002}, and walking \cite{winter_biomechanical_1983}). In those studies, the joint torques are normalized by body mass; in \cite{roebroeck_biomechanics_1994}, they are also normalized by body height. In this section, we assume a user mass of \SI{67.8}{\kilo\gram} and height of \SI{1.76}{\meter}, which correspond to the average mass and height of the participants in \cite{roebroeck_biomechanics_1994} (6 females and 4 males).
\subsubsection{SOMA vs. SOSA}
Each of the trajectories from inverse kinematics and dynamics defines a matrix $\bs{Q}$ and a vector $\bs{q}$, which are the parameters for the optimization problem (\ref{eq:SOMA}). In the SOMA, the sit-to-stand trajectories define the parameters in the optimization objective, and the rest of the trajectories define the parameters of the quadratic constraints. The SOSA uses the same optimization objective as the SOMA without the quadratic constraints. The SOMA reduced \kneeSOMATorquePercent{} sit-to-stand RMS-knee-torque compared to not wearing the parallel spring. The SOSA reduced it by \kneeSOSATorquePercent{}. The main advantage of the SOMA over the SOSA is that it guarantees that RMS torque will not increase for any other tasks (\eg, walking), as shown in Fig.~\ref{fig:torqueResultsKneeExo}. In terms of optimal stiffness and preload, the SOMA represents the point within the intersection of all the ellipses that minimizes sit-to-stand RMS torque. The SOSA is the center of the sit-to-stand ellipse, \ie, the point that globally minimizes sit-to-stand RMS-knee-torque (Fig.~\ref{fig:ellipseResultsKneeExo}).

The ellipses (Fig.~\ref{fig:ellipseResultsKneeExo}) and numerical solver results (Fig.~\ref{fig:torqueResultsKneeExo}) complement each other. Expressing beneficial spring parameters as points within a ellipse contextualizes the selection of stiffness and preload with respect to a set of tasks. For example, the relative size of the stair descent and walking ellipses justifies prioritizing walking -- any design that benefits walking also benefits stair ascent. Numerical optimization complements the geometric approach for a large number of tasks. For example, a design that includes walking at a continuous range of speeds and inclines may represent an intractable number of ellipses. In this scenario, the optimization solver can provide a single point that makes the right compromise between the large set of tasks.

\begin{figure}
    \begin{tikzpicture}
\begin{axis}[
	width = 0.5\textwidth,
	height = 0.22\textheight,
	ybar = 2pt,
	axis x line = bottom,
	x axis line style = {-},
	axis y line = left,
	y axis line style={opacity=0},
	ytick=\empty,
	enlargelimits = 0.16,
	legend style={
		 at={(0.3,0.95)},
		draw = none,
		legend cell align = left,
	},
	ylabel={RMS Torque \lbrack\SI{}{\newton\meter}\rbrack},
	symbolic x coords={Walk, Stair Ascend, Stair Descend, Sit to Stand},
	x tick label style={
		text width=1.5cm,
		align=center,
	},
	bar width=13pt,
	xtick=data,
	nodes near coords = {\scriptsize \pgfmathprintnumber{\pgfplotspointmeta}},
	nodes near coords align={vertical},
	]
	\addplot [blue!100, fill=blue!20] coordinates {(Walk, 16.1)(Stair Ascend, 28.3)(Stair Descend, 44.9)(Sit to Stand, 27.7)};
	\addplot [black!100, fill=gray!20, postaction={pattern=north east lines}] coordinates {(Walk, 16.1)(Stair Ascend, 26.8)(Stair Descend, 39.6)(Sit to Stand, 21.0)};
	\addplot [red!100, fill=red!20] coordinates {(Walk, 19.8)(Stair Ascend, 26.9)(Stair Descend, 36.5)(Sit to Stand, 20.5)};
	\legend{No spring, SOMA, SOSA}
	\end{axis}
\end{tikzpicture} 
    \caption{RMS knee torque across multiple activities with a parallel spring optimized for multiple activities (SOMA) and a spring optimized for a single activity (SOSA). The stiffness and preload for each spring are in Fig.~\ref{fig:ellipseResultsKneeExo}.}
    \label{fig:torqueResultsKneeExo}
\end{figure}
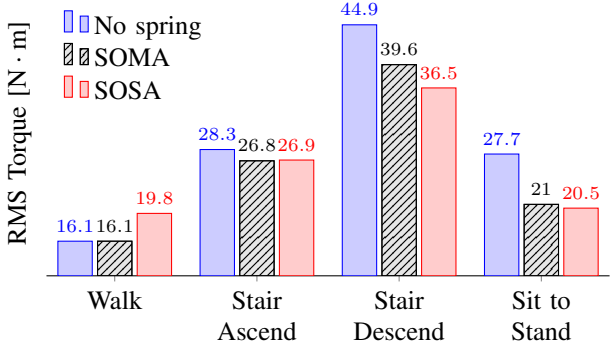
\begin{figure}
    \centering
    \input{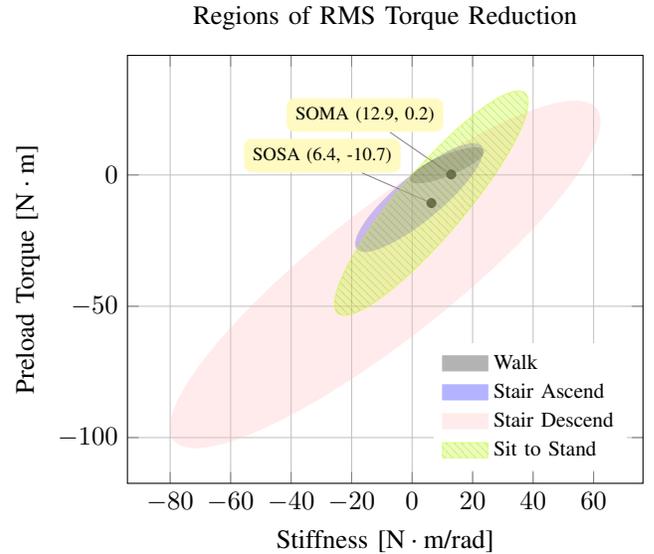}
    \caption{Geometric approach to select the stiffness and preload of the spring in parallel with a biological knee that minimizes its RMS torque during sit-to-stand transitions. Any point in each ellipse interior represents a combination of parallel stiffness and preload that reduces RMS torque compared to a no-spring configuration. Each ellipse center denotes the optimal spring for an individual task without any consideration of other tasks (SOSA). The SOMA denotes the point that minimizes sit-to-stand RMS knee torques without increasing it for other tasks. The SOMA (SOSA) stiffness and preload are \kneeSOMAstiffness{} (\kneeSOSAstiffness{}) and \kneeSOMApreload{} (\kneeSOSApreload{}). For a detailed discussion on the trade-off analysis please see Section~\ref{sec:ObjectiveFunctionsAndTradeOff}}
    \label{fig:ellipseResultsKneeExo}
\end{figure}

\section{Discussion} \label{sec:discussion}
\subsection{Objective function and constraints selection}\label{sec:ObjectiveFunctionsAndTradeOff}
As demonstrated in our prosthesis and exoskeleton applications, the convex quadratic functions can define the optimization constraints or the objective function (\eg, in a weighted sum). Including a task as a constraint or part of the objective depends on the application and best judgment of the designer. Using the task as a constraint, the optimal design will not ``hurt" a specific task; however, the optimal design may be on the ellipse boundary, which implies no benefit for the specific task (\eg, SOMA in Fig.~\ref{fig:ellipseResultsKneeExo}). The right-hand side of each inequality can include an offset so that the optimal point is in the ellipse interior, \ie, reduces the cost for a task. This offset can lead to infeasible results. Using the task in the objective, the optimal design can be in the interior of a given ellipse, but it requires adequate selection of the objective function to guarantee a benefit for multiple tasks. For example, our SOMA and SOSA in Fig.~\ref{fig:ellipseResultsAnkleProstheses} are in the interior of all the ellipses except for sit-to-stand, explaining the larger percentage in energy consumption for sit-to-stand (Table~\ref{table:ExperimentalResults}). In some applications, it can be appropriate to increase the cost for a set of tasks to minimize the overall cost for all tasks.

The trade-off between single- and multi-task optimality (SOSA vs. SOMA) depends on the kinematics and kinetics for each task. The orientation and boundaries of the corresponding ellipses can guide the trade-off analysis. In terms of orientation, the semi-minor axis for a given ellipse is the direction where changes in stiffness and preload represent a higher increase in the cost (\ie, RMS effort or energy consumption). In terms of boundaries, the ellipse boundary is the set of all stiffnesses and preloads that do not increase the cost with respect to a no-spring condition. These two principles can help interpret the ellipse plot. For example, in Fig.~\ref{fig:ellipseResultsAnkleProstheses}, the SOMA is outside the boundary of the Sit to Stand ellipse and between the semi-minor and -major axes. Thus, the designer should anticipate a mid-to-high cost in energy consumption for Sit to Stand. Note that Walking is the activity with the highest relative weight in the objective function (2000/2280). Thus, given that the semi-major axis in the walking ellipse is in a similar orientation as the stiffness axis, it is reasonable to reduce the stiffness of the SOSA (moving in the x-axis towards the other ellipse centers) to account for multiple activities. This analysis aims to provide an intuitive interpretation to our geometric approach. Our recommendation is to select target design parameters from the numerical approach. In addition to the analysis of ellipse boundaries and orientations, the tradeoff analysis can use any of the multi-objective optimization techniques in numerical optimization (\eg, Pareto front) \cite{boyd_convex_2004}.

\subsection{Combining the optimization and geometric approach}
Combining the graphical and optimization approaches takes advantage of their complementary strengths: visual intuition from the intersection of ellipses and scalability from the numerical optimization. Evaluating the benefit of a parallel spring as the intersection of ellipses guides the design process for constraints not specified in this work, \eg,  the spring stiffness can only change based on a discrete set of options specified by a manufacturer. The numerical optimization approach accurately finds designs that guarantee performance for an arbitrary set of tasks. For the sake of analysis, any actuator is a PEA---the parallel spring stiffness and preload can be zero to represent the case when there is no spring (\ie, $\bs{x}=0$). From this perspective, the more generic the set of tasks the less stiffness and preload is expected from the optimal parallel spring design. This perspective agrees with Brown and Ulsoy~\cite{robert_brown_maneuver_2013} where, for an arbitrarily large set of tasks, the optimal spring stiffness is zero. Our ellipses also capture this intuition as the point $\bs{x}=0$ always satisfies the inequalities in (\ref{eq:QCQP}). Thus, $\bs{x}=0$ is always an element at the intersection of all the ellipses. Our recommendation is for designers to use the software \cite{source_code} that supplements this work to plot the ellipses and numerically solve the optimization problem. The goal of this software is for the reader to use the proposed optimization and geometric approach without investing significant time in programming it.
\subsection{Applications in variable stiffness actuators}
The convexity of our optimization problem enables numerical solutions that are fast (in polynomial time), independent of solver initial conditions, and have guaranteed global optimality. Mechanisms that change spring stiffness (\eg, \cite{mathews_design_2022,grioli_variable_2015,groothuis_variable_2014}) can benefit from our approach as our convex optimization approach is suitable for real-time solutions.
\subsection{Limitations}
The geometric and optimization methods depend on knowledge of the kinematics and kinetics of the load. Those requirements may be unknown or, at the very least, variable. Our method pairs well with simulations, datasets, or existing recordings of the joint effort and flow. However, if the load kinematics and kinetics are not available, our approach will not apply. In addition, our methods depend on the time scale for each activity. For example, we set the weights in the objective for the prosthetic design to be proportional to the use as reported in the existing literature. This time scale may be difficult to approximate at the design stage. A parallel spring can be interpreted as a compliant mechanism in parallel with the effort source. The design of a compliant mechanism depends on three factors: the load-elongation profile (\eg, stiffness and preload), the material properties (\eg, Young's modulus), and geometry (\eg, spiral torsional spring) \cite{howell_compliant_2001,guerinot_compliant_2005,howell_handbook_2013,peterson_cross-axis_2024}. This paper focuses on the load-elongation profile. Future work will focus on the material properties and geometry. Future work will also consider the optimization of a nonlinear load-elongation profile, as the current approach is limited to linear springs with non-zero preload.
\section{Conclusion}\label{sec:conclusion}
This paper presents a framework to optimize the stiffness and preload of a parallel spring that minimizes the energy consumption and RMS effort for an arbitrary effort source that actuates multiple tasks. The optimization problem is a convex quadratically constrained quadratic program that can be geometrically interpreted as the intersection of ellipses in the stiffness and preload plane. Traditional optimization techniques do not guarantee benefits for an arbitrarily large set of activities. The main advantage of the proposed approach is to optimize stiffness and preload values that guarantee the benefits of a parallel spring (minimizing RMS torque or energy consumption) for an arbitrarily large set of activities (\eg, walking at multiple speeds, stair ascent/descent with arbitrary stair inclines). Identifying ellipses in the stiffness and preload plane that guarantee spring benefits is also advantageous to select spring parameters. Defining regions of benefits instead of a singular point (traditional output of numerical optimizers) enables intuition and flexibility in the design to accommodate additional implementation constraints, such as limited manufacturing accuracy. To ease adoption, the reader is encouraged to test the framework using the source code that supplements this work (\verb|https://werolab.nd.edu/psea|) \cite{source_code}. Readers can use the code to investigate the impact of varying reference trajectories (\eg, multiple walking speeds) on the optimization results. We applied our framework to the design of a knee exoskeleton and a powered prosthetic ankle. The example code for the two applications is also in \cite{source_code}. 
%

\begin{IEEEbiography}[{\includegraphics[width=1in,height=1in,clip,keepaspectratio]{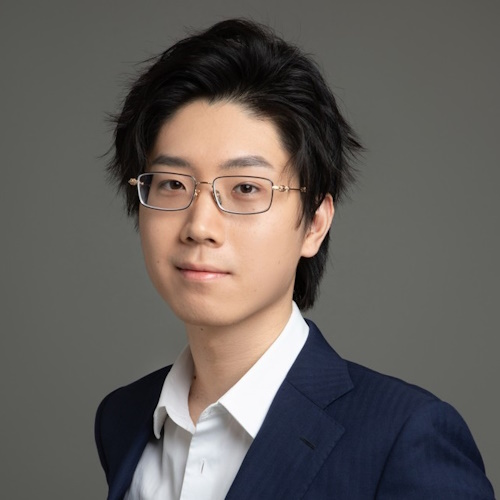}}]{Kang Yang}
(Graduate Student Member, IEEE) was born in Shanghai, China. He received the B.S. degree in mechanical engineering from Columbia University, New York City, NY, USA, in 2020, the M.S. degree in mechanical engineering from the University of Michigan, Ann Arbor, MI, USA, in 2022, and is currently pursuing the Ph.D. degree in mechanical engineering at the University of Notre Dame, Notre Dame, IN, USA.

From 2020 to 2022, he was with the Precision Systems Design Laboratory at the University of Michigan, where he designed and manufactured a 3 DoF tendon-driven index finger and thumb, and developed a compact prosthetic wrist. Since 2022, he has been with the Wearable Robotics Laboratory at the University of Notre Dame, focusing on thermally and energetically optimal actuator design, and solving convex optimization problems for prosthetic applications. He is the author of several published articles in journals such as IEEE Transactions on Medical Robotics and Bionics and has presented his research at international conferences, including the IEEE/ASME International Conference on Advanced Intelligent Mechatronics.
\end{IEEEbiography}

\begin{IEEEbiography}[{\includegraphics[width=1in,height=1in,clip,keepaspectratio]{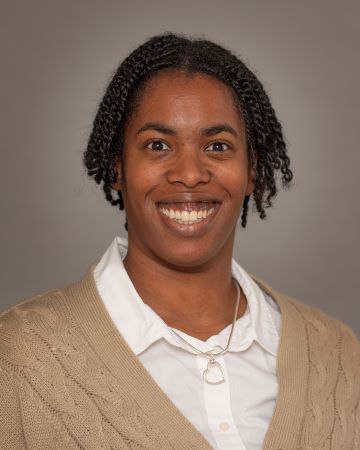}}]{Myia M. Dickens}
(Graduate Student Member, IEEE) received the B.S. degree in mechanical engineering from the University of California, Irvine, CA, USA, in 2021.

From 2019 to 2021, she was a Prototype Lab Intern at UCI Beall Applied Innovation. In 2021, she was a Summer Intern at Idaho National Laboratory in the Material Science and Engineering Department. In 2023, she was a Summer Intern at General Motors in the Battery Systems and Architecture Department. She is currently a Ph.D. Student in the Department of Aerospace and Mechanical Engineering at the University of Notre Dame, Notre Dame, IN, USA. Her research has been concerned with robot locomotion and powered lower-limb prostheses.

Ms. Dickens is a Graduate Student Member of ASME and a Collegiate Member of SWE and NSBE.
\end{IEEEbiography}

\begin{IEEEbiography}[{\includegraphics[width=1in,height=1in,clip,keepaspectratio]{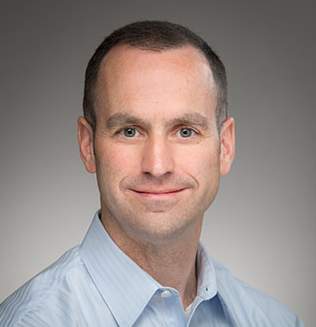}}]{James P. Schmiedeler}
(Senior Member, IEEE) received the B.S. degree in mechanical engineering from the University of Notre Dame, Notre Dame, IN, USA, in 1996 and the M.S. and Ph.D. degrees in mechanical engineering from The Ohio State University, Columbus, OH, USA in 1998 and 2001, respectively. 

From 2002 to 2003, he was an Assistant Professor in the Department of Mechanical and Industrial Engineering at the University of Iowa. From 2003 to 2008, he was an Assistant Professor in the Department of Mechanical Engineering at The Ohio State University. He is currently a Professor in the Department of Aerospace and Mechanical Engineering at the University of Notre Dame, Notre Dame, IN, USA. His research has been concerned with biped robot locomotion, powered lower-limb prostheses, and rehabilitation of human walking.

Prof. Schmiedeler is a Fellow of ASME.
\end{IEEEbiography}

\begin{IEEEbiography}[{\includegraphics[width=1in,height=1in,clip,keepaspectratio]{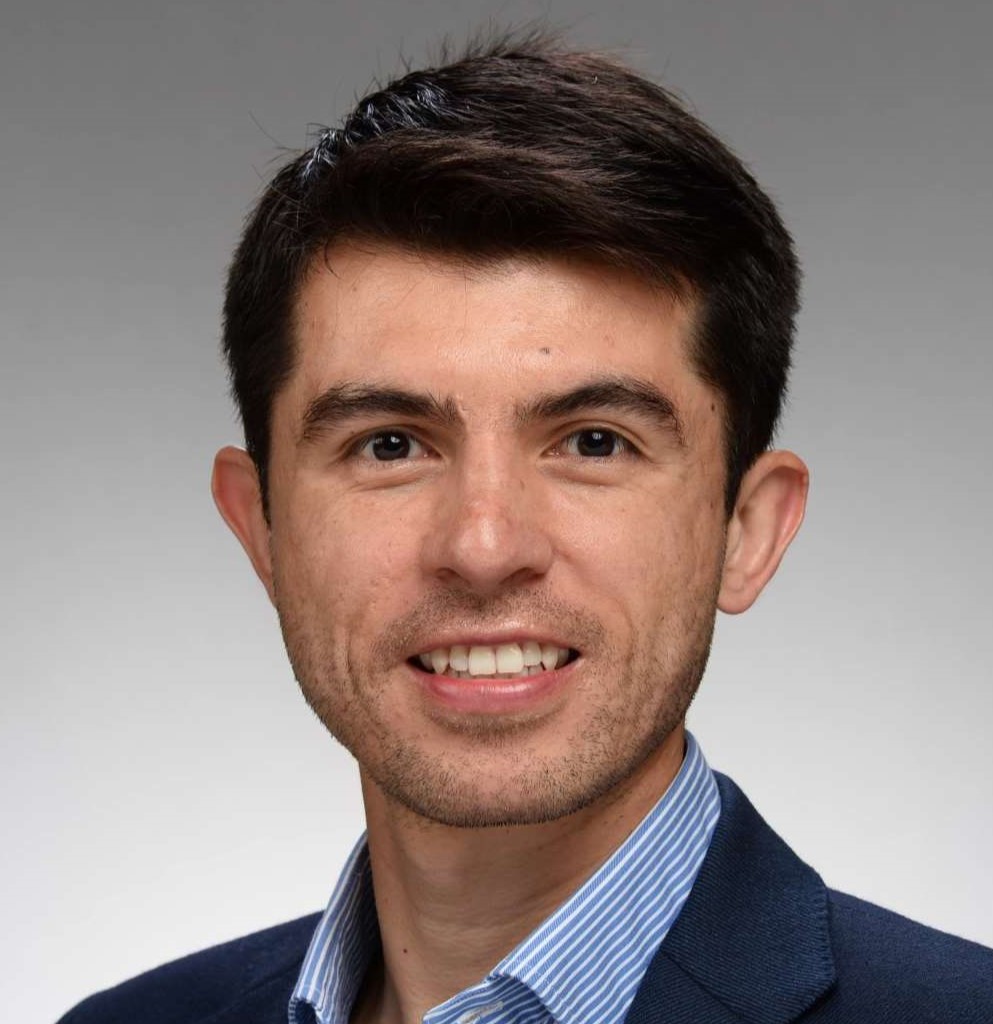}}]{Edgar Bol\'ivar-Nieto}
(Member, IEEE) is an Assistant Professor in the Department of Aerospace and Mechanical Engineering at the University of Notre Dame, Notre Dame, IN, USA. Dr. Bolívar-Nieto earned his Ph.D. and M.S. degrees in mechanical engineering from The University of Texas at Dallas, Richardson, TX, USA; and his B.S. degree in mechatronics engineering from the Universidad Nacional de Colombia, Bogot\'a D.C., Colombia. From 2019 to 2021, he was a Research Fellow at the University of Michigan, Ann Arbor, MI, USA, where he contributed to the control of lower-limb-powered prostheses.  In November 2021, Dr. Bolívar-Nieto started as an Assistant Professor and founded the Wearable Robotics Laboratory at the University of Notre Dame. The laboratory focuses on the application of real-time and robust optimization algorithms for the mechanical design, control, and estimation of wearable robots. Currently, Dr. Bolívar-Nieto serves as Associate Editor in the journals Scientific Data and IEEE Transactions on Neural Systems and Rehabilitation Engineering.
\end{IEEEbiography}

\end{document}